\documentclass{article}

\usepackage{arxiv}

\usepackage[utf8]{inputenc} 
\usepackage[T1]{fontenc}    
\usepackage{hyperref}       
\usepackage{url}            
\usepackage{booktabs}       
\usepackage{amsfonts}       
\usepackage{nicefrac}       
\usepackage{microtype}      
\usepackage{cleveref}       
\usepackage{lipsum}         
\usepackage{graphicx}
\usepackage{natbib}
\usepackage{doi}
\usepackage[title]{appendix}%
\usepackage{xspace}
\usepackage{multirow}
\usepackage{lscape}
\usepackage{longtable}
\usepackage{listings}
\usepackage{xcolor}
\usepackage{nameref}

\definecolor{codegreen}{rgb}{0,0.6,0}
\definecolor{codegray}{rgb}{0.5,0.5,0.5}
\definecolor{codepurple}{rgb}{0.58,0,0.82}
\definecolor{backcolour}{rgb}{0.95,0.95,0.92}

\date{}

\hypersetup{
    pdftitle={Learning to Solve Abstract Reasoning Problems with Neurosymbolic Program Synthesis and Task Generation},
    pdfauthor={Jakub Bednarek, Krzysztof Krawiec},
    pdfkeywords={Neurosymbolic systems, Program synthesis, Abstract reasoning},
}

\newcommand{\mnamens}{TransCoder} 
\newcommand{\mname}{\mnamens\xspace}

\title{Learning to Solve Abstract Reasoning Problems with Neurosymbolic Program Synthesis and Task Generation}

\author{
        \href{https://orcid.org/0000-0003-0727-3442}
        {\includegraphics[scale=0.06]{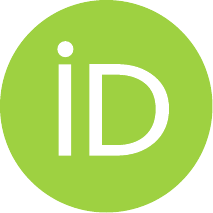}\hspace{1mm}Jakub Bednarek}
        \thanks{Alternative email: \texttt{jakub.bednarek.g@gmail.com}}\\
	Institute of Computing Science\\
	Poznan University of Technology\\
	Poland \\
	\texttt{jakub.bednarek@put.poznan.pl} \\
	\And
	\href{https://orcid.org/0000-0001-5439-3231}
        {\includegraphics[scale=0.06]{orcid.pdf}\hspace{1mm}Krzysztof Krawiec} \\
	Institute of Computing Science\\
	Poznan University of Technology\\
	Poland \\
	\texttt{krzysztof.krawiec@cs.put.poznan.pl} \\
}

\renewcommand{\shorttitle}{Learning from Mistakes with \mname}

\begin{document}
\maketitle

\begin{abstract}
The ability to think abstractly and reason by analogy is a prerequisite to rapidly adapt to new conditions, tackle newly encountered problems by decomposing them,  and synthesize knowledge to solve problems comprehensively. We present \mname, a method for solving abstract problems based on neural program synthesis, and conduct a comprehensive analysis of decisions made by the generative module of the proposed architecture. At the core of \mname is a typed domain-specific language, designed to facilitate feature engineering and abstract reasoning. In training, we use the programs that failed to solve tasks to generate new tasks and gather them in a synthetic dataset. As each synthetic task created in this way has a known associated program (solution), the model is trained on them in supervised mode. Solutions are represented in a transparent programmatic form, which can be inspected and verified. We demonstrate \mname's performance using the Abstract Reasoning Corpus dataset, for which our framework generates tens of thousands of synthetic problems with corresponding solutions and facilitates systematic progress in learning.
\end{abstract}


\section{Introduction}
Abstract reasoning tasks have a long-standing tradition in AI (e.g. Bongard problems \cite{bongard1967problem}, Hofstadter's analogies \cite{Hofstader1995}). In the past, they have been most often approached with algorithms relying exclusively on symbolic representations and typically involving some form of principled logic-based inference. While this can be successful for problems posed `natively' in symbolic terms (e.g. \cite{Hofstader1995}), challenges start to mount up when a symbolic representation needs to be inferred from a low-level, e.g. visual, representation \cite{bongard1967problem}. The recent advances in deep learning and increasing possibilities of their hybridization with symbolic reasoning (see Sec.\ \ref{sec:related}) opened the door to architectures that combine `subsymbolic' processing required to perceive the task with sound symbolic inference.  

This study introduces \mname, a neurosymbolic architecture that relies on programmatic representations to detect and capture relevant patterns in low-level representation of the task, infer higher-order structures from them, and encode the transformations required to solve the task. \mname is designed to handle the tasks from the Abstract Reasoning Corpus (ARC,  \cite{CholletARC}), a popular benchmark that epitomizes the above-mentioned challenges. Our main contributions include (i) the original neural architecture that synthesizes programs that are syntactically correct by construction, (ii) the `learning from mistakes' paradigm to provide itself with a learning gradient by synthesizing tasks of adequate difficulty, (iii) an advanced perception mechanism to reason about small-size rasters of variable size, and (iv) empirical assessment on the ARC suite. 


\section{Abstract Reasoning Corpus}\label{sec:arc}

\begin{figure}[t!]
    \centering
    \includegraphics[width=1\linewidth]{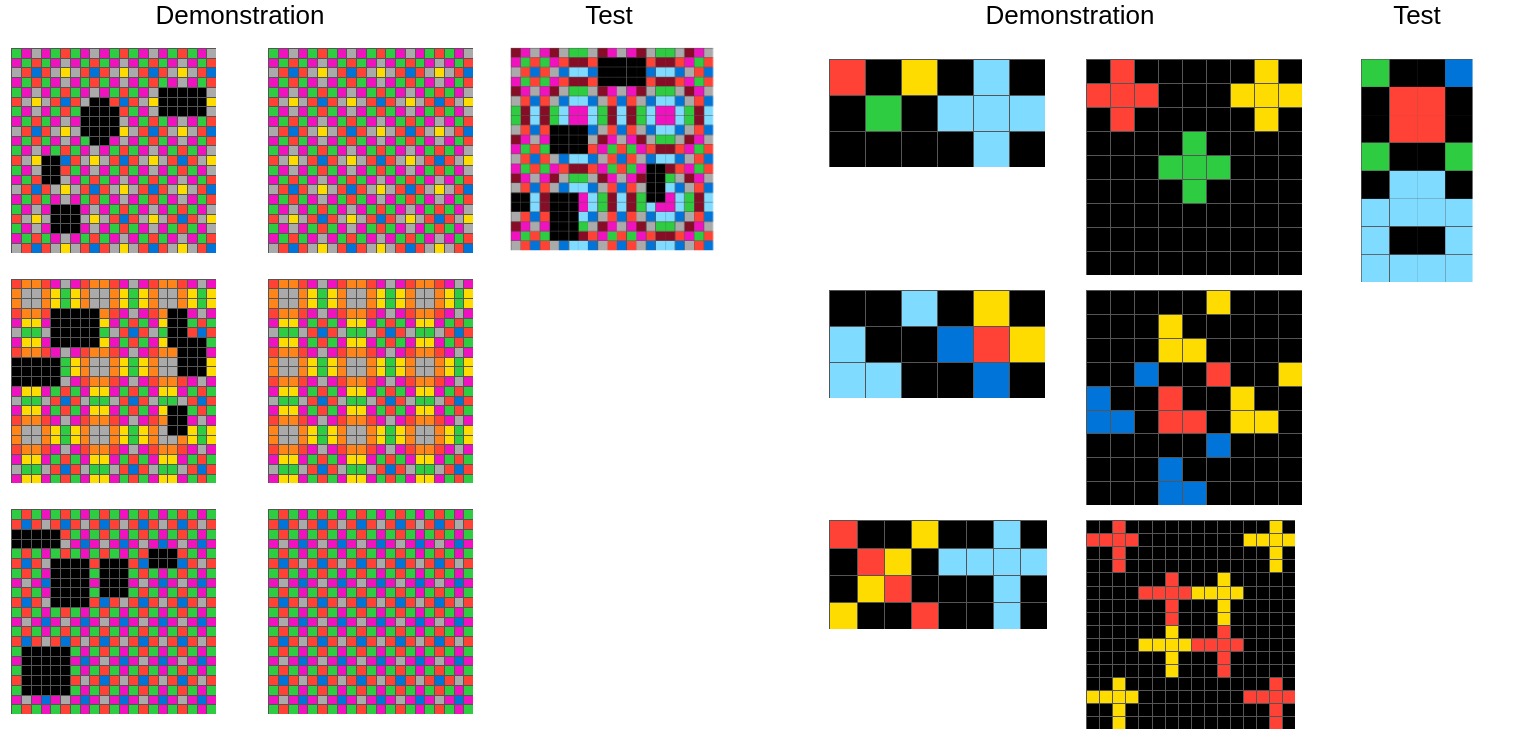}
    \caption{Examples from the Abstract Reasoning Corpus Dataset.}
    \label{fig:task}
\vspace{-5mm}
\end{figure}

Abstract Reasoning Corpus (ARC) \cite{CholletARC} is a collection of 800 visual tasks, partitioned into 400 training tasks and 400 testing tasks.\footnote{\url{https://github.com/fchollet/ARC}} Each task comprises a few (usually 3, maximally 6) \emph{demonstrations} and a \emph{test} (Fig.\ \ref{fig:task}). A demonstration is a pair of raster images, an input image and an output image. Images are usually small (at most 30 by 30 pixels) and each pixel can assume one of 10 color values, represented as a categorical variable (there is no implicit ordering of colors). The test is also a raster image, meant to be interpreted as yet another input for which the corresponding output is unknown to the solver. 

For each ARC task, there exists a unique processing rule (unknown to the solver) that maps the input raster of each demonstration to the corresponding output raster. The solver is expected to infer\footnote{Or, more accurately, \emph{induce}, as the demonstrations never exhaust all possible inputs and outputs.} that rule from the demonstrations and apply it to the test raster to produce the corresponding output. The output is then submitted to the oracle which returns a binary response informing about the correctness/incorrectness of this solution. 

The ARC collection is very heterogeneous in difficulty and nature, featuring tasks that range from simple pixel-wise image processing, re-coloring of objects, to mirroring of the parts of the image, to combinatorial aspects (e.g. counting objects), to intuitive physics (e.g. an input raster to be interpreted as a snapshot of moving objects and the corresponding output presenting the next state). In quite many tasks, the black color should be interpreted as the background on which objects are presented; however, there are also tasks with rasters filled with `mosaics' of pixels, with no clear foreground-background separation (see e.g. the left example in Fig.\ \ref{fig:task}). Raster sizes can vary between demonstrations, and between the inputs and outputs; in some tasks, it is the \emph{size} of the output raster that conveys the response to the input. Because of these and other characteristics, ARC is widely considered extremely hard: in the Kaggle contest accompanying the publication of this benchmark\footnote{\url{https://www.kaggle.com/c/abstraction-and-reasoning-challenge}}, which closed on the 28th of May 2020, the best contestant entry algorithm achieved an error rate of 0.794, i.e. solved approximately 20\% of the tasks from the (unpublished) evaluation set, and most entries relied on a computationally intensive search of possible input-output mappings.


\section{The proposed approach}\label{sec:approach}

\begin{figure}[t!]
    \centering
    \includegraphics[width=1\linewidth]{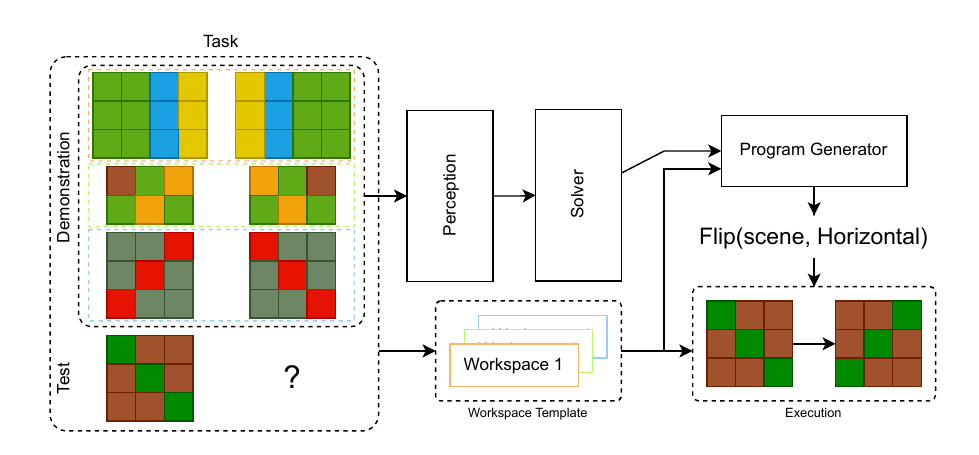}
    \caption{The overall architecture of \mname. 
    }
    \label{fig:architecture}
\end{figure}

The broad scope of visual features, object properties, alternative interpretations of images, and inference mechanisms required to solve ARC tasks suggest that devising a successful solver requires at least some degree of symbolic processing. It is also clear that reasoning needs to be \emph{compositional}; e.g. in some tasks objects must be first delineated from the background and then counted, while in others objects need to be first counted, and only then the foreground-background distinction becomes possible. It is thus essential to equip the solver with the capacity to rearrange the inference steps in an (almost) arbitrary fashion. 

The above observation is a strong argument for representing the candidate solutions as \emph{programs} and forms our main motivation for founding \mname on \emph{program synthesis}, where the solver can express a candidate solution to the task as a program in a \emph{Domain-Specific Language} (DSL), a bespoke programming language designed to handle the relevant entities. Because (i) the candidate programs are to be generated in response to the content of the (highly visual) task, and (ii) it is desirable to make our architecture efficiently trainable with gradient to the greatest degree possible, it becomes natural to control the synthesis using a neural model. Based on these premises, \mname is a \emph{neurosymbolic system} that comprises (Fig.\ \ref{fig:architecture}): 
\begin{itemize}
    \item \textbf{Perception module}, a neural network that maps demonstrations to a latent vector $z$ of fixed dimensionality,  
    \item \textbf{Solver}, a (stochastic) network that maps the latent representation of the task $z$ to the latent representation $z'$ of the to-be-synthesized program,
    \item \textbf{Program generator}, (\textbf{Generator} for short) a recurrent network that maps $z'$ to the program $p$ represented as an Abstract Syntax Tree,
    \item \textbf{Program interpreter}, (\textbf{Interpreter} for short) which executes $p$, i.e. applies it to rasters.  
\end{itemize}
In training, the Interpreter applies $p$ independently to each of the input rasters of demonstrations and returns the predicted output rasters, which are then confronted with the true output rasters using a loss function. 
In testing, $p$ is applied to the test raster and the resulting raster is submitted to the oracle that determines its correctness.

We detail the components of \mname in the following sections. For technical details, see Appendix A.

\subsection{The Perception Module}\label{sec:perception}

\begin{figure}[t!]
    \centering
    \includegraphics[width=1\linewidth]{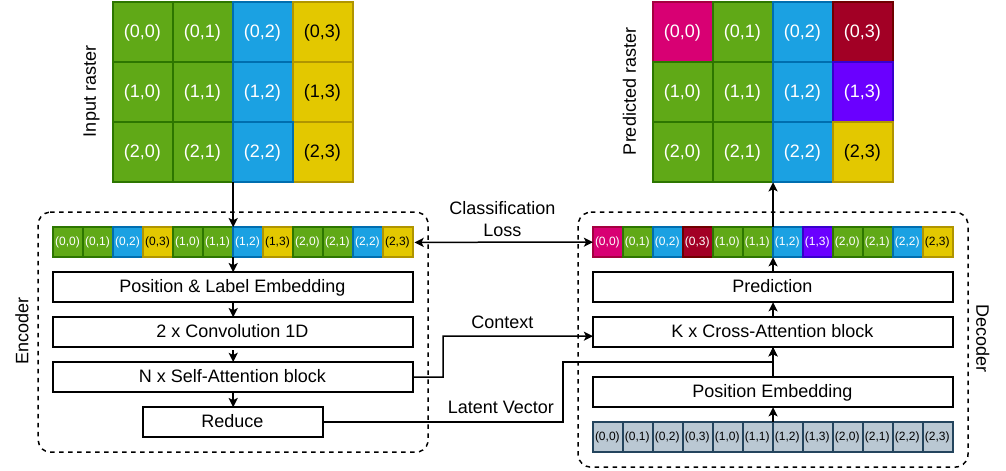}
    \caption{The autoencoder architecture used to pre-train the raster encoder in Perception. Left: the encoder (used in \mname after pre-training). Right: the decoder (used only in pre-training and then discarded).}
    \label{fig:autoencoder}
\end{figure}

The perception module comprises the \emph{raster encoder} and the \emph{demonstration encoder}. 

The \textbf{raster encoder} is based on an attention module that allows processing rasters of different sizes, which is required when solving ARC tasks (see, e.g., the task shown in Fig.\ \ref{fig:architecture}). However, raster sizes need to be taken into account, as they often convey crucial information about the task. To convey it to the model, we tag each pixel with the color \emph{and} its $(x,y)$ coordinates, the latter acting as a \emph{positional embedding}. The image is flattened to a tagged sequence of tokens representing pixels (see the left part of Fig.\ \ref{fig:autoencoder}). The tensor resulting from the Reduce block forms the fixed-length representation of the raster, subsequently fed into the demonstration encoder. 

The raster encoder is pre-trained within an autoencoder framework, where the raster encoder is combined with a compatible decoder that can reproduce varying-length sequences of pixels (and thus the input raster) from a fixed-dimensionality latent (the right part of Fig.\ \ref{fig:autoencoder}). The pre-training is intended to make feature extraction invariant to permutations of colors, while accounting for the black color serving the special role of background in a large share of tasks. To this end, the rasters derived from the ARC are treated as dynamic templates colored on the fly during training while preserving the distinctions of colors. As a result of pre-training, the raster encoder achieves the per-pixel reconstruction accuracy of 99.98\% and the per-raster accuracy of 96.36\% on the testing part of the original ARC.

The \textbf{demonstration encoder} concatenates the latent vectors obtained from the raster encoder for the input and output raster of a demonstration and passes them through a two-layer MLP. This is repeated for all demonstrations, and the output vectors produced by the MLP are chained into a sequence, which is then subject to processing with four consecutive self-attention blocks. The sequence of vectors produced in this process is averaged, resulting in a fixed-dimensionality (independent of the number of demonstrations) vector $z$ representing a task.

\subsection{Solver}\label{sec:solver}

The latent $z$ produced by Perception forms a compressed representation of raster images in an ARC task. As it has been almost perfectly pre-trained via auto-association (see previous section), we expect it to contain the entirety of information about the \emph{content} of the input raster. However, this does not mean that it conveys the knowledge sufficient for \emph{solving} the task. The role of the Solver is to map $z$ to a latent representation $z'$ of the program to be generated. Technically, Solver is implemented as a two-layer MLP.

However, as in most programming languages, the relationship between the programs written in our DSL and their input-output behaviors is many-to-one, i.e. the same mapping from the input to output raster can be implemented with more than one program. As a result, the relationship between DSL programs and ARC \emph{tasks} is many-to-many, i.e. a given task can be solved with more than one DSL program, and this very program can be a solution to more than one ARC task. 

To account for this absence of one-to-one correspondence, we make the Solver stochastic by following the blueprint of the Variational Autoencoder (VAE,~\cite{KingmaVAE}): the last layer of the Solver does not directly produce $z'$, but parameterizes the normal distribution with two vectors $z_\mu$ and $z_\sigma$ and then calculates $z' = z_\mu + z_\sigma N(0,1)$ where $N(0,1)$ is generated by the random number generator. The intent is to allow $z_\mu$ to represent the centroid of the set of programs that solve a given task, while $z_\sigma$ to model the extent of that set in the latent space.

\subsection{Workspace}

As many ARC tasks involve qualitative and combinatorial concepts (e.g. counting objects), we supplement the information provided by Perception with selected symbolic percepts inferred from the task independently from Perception, via direct `procedural parsing' of images. We provide two data structures for that purpose: 
\begin{itemize}
    \item  \emph{Workspace Template} that contains the abstract placeholders for entities that appear in \emph{all} demonstrations of a given task, 
    \item \emph{Workspaces} that `instantiate` that template with concrete values derived from particular demonstrations. 

\end{itemize}

The entries in a Workspace Template act as \emph{keys} that index specific \emph{values} (realizations) in the Workspace of a specific demonstration. For instance, the \emph{Scene} symbol in the Workspace Template may have a different value in each Workspace. For the task shown in Fig.~\ref{fig:architecture}, the Workspace Template contains the \emph{Scene} symbol, and its values in the first two Workspaces are different:

\begin{lstlisting}[language=Python,xleftmargin=0.05\linewidth]
scene_0 = Region(  # first demonstration
    positions=[[0,0], [0,1], [0,2], ...], 
    colors=[Green, Green, Blue, ...]
)
scene_1 = Region(  # second demonstration
    positions=[[0,0], [0,1], [0,2], ...], 
    colors=[Brown, Green, Orange, ...]
)
\end{lstlisting}

The list of workspace keys is predefined and includes constants (universal symbols shared between all tasks), local invariants (values that repeat within a given task, e.g. in each input raster), and local symbols (information specific to a single demonstration pair, e.g. an input \emph{Region}). 
For a complete list of available keys, see~\nameref{app:dsl}.

The workspaces require appropriate `neural presentation' for the Generator of DSL programs. For a given task, all symbols available in the Workspace Template are first embedded in a Cartesian space, using a learnable embedding similar to those used in conventional DL. This representation is \emph{context-free}, i.e. each symbol is processed independently. We then enrich this embedding with the information in the latent $z'$ produced by the Solver (Sec.~\ref{sec:solver}) by concatenating both vectors and processing them with a two-layer MLP, resulting in a \emph{contextual embedding} of the symbol. 

Moreover, the DSL's operations are also included in the Workspace; each of them is also embedded in the same Cartesian space so that the Generator can choose from them alongside the symbols from the workspace.
In this way, the elements of the DSL are presented to the Generator (on the neural level) in the context of the given task, and symbols (such as `red`) may have a different embedding depending on the perception result. This is expected to facilitate alternative interpretations of the roles of particular percepts; for instance, while the black pixels should often be interpreted as the background, some tasks are exceptions to this rule.

\subsection{The Domain-Specific Language}

The DSL we devised for \mname is a typed, functional programming language, with leaves of AST trees fetching input data and constants, and the root of the tree producing the return value. Each operation (an inner AST node) is a function with a typed signature and implementation. The DSL features concrete (e.g. \emph{Int}, \emph{Bool}, or \emph{Region}) and generic (e.g. \emph{List[T]}) data types.

A complete DSL program has the signature \emph{Region} $\rightarrow$ \emph{Region}, i.e. it can be applied to the input of an ARC demonstration (or the query) and produce the corresponding output. 

The current version of the DSL contains 40 operations, which can be divided into data-composing operations (form a more complex data structure from constituents, e.g. \emph{Pair}, \emph{Rect}), property-retrieving operations (fetch elements or extract simple characteristics from data structures, e.g. \emph{Width}, \emph{Area} or \emph{Length}), data structure manipulations (e.g. \emph{Head} of the list, \emph{First} of a \emph{Pair}, etc.), arithmetics (\emph{Add}, \emph{Sub}, etc.), and region-specific operations (high-level transformations of drawable objects, e.g. \emph{Shift}, \emph{Paint}, \emph{FloodFill}). Our DSL features also higher-order functions known from functional programming, for example, \emph{Map} and \emph{Filter} which apply an argument in the form of a subprogram to elements of a compound data structure like a \emph{List}. The complete definition of the DSL can be found in~\nameref{app:dsl}.

\subsection{Program Generator}

The Program Generator (Generator for short) is a bespoke architecture based on the blueprint of the doubly-recurrent neural network (see~\cite{AlvarezDRNN} for a simple variant of DRNN). The latent $z'$ obtained from the Solver becomes the initial state of this network, which then iterates over the nodes of the Abstract Syntax Tree (AST) of the program being generated in the breadth-first order. For the root node of the AST, the return type is \emph{Region} for the root node; for other nodes, it is determined recursively by the types of arguments required by DSL functions picked in previous iterations. 

In each iteration, the Generator receives the data on the current context of AST generation, including the current size (the number of already generated nodes in the AST) and depth of the node in the AST, the parent of the current node, and the return type of the node. It is also fed with the set of symbols available in the workspaces (including the elements of the DSL), via the embedding described in the previous section. From this set, the Generator selects the symbols that meet the requirements regarding the type and the maximum depth of the tree. Then, it applies an attention mechanism to the embedded representations of the selected symbols. The outcome of attention is the symbol to be `plugged' into the AST at the current location. 

The Generator also determines the hidden state of the DRNN to be passed to each of the child nodes. This is achieved by merging the current state with a learnable embedding indexed with children's indices, so that generation in deeper layers of the AST tree is informed about node's position in the sequence of parent's children. The generation process iterates recursively until the current node requests no children, which terminates the current branch of the AST (but not the others). It is also possible to enforce termination by narrowing down the set of available symbols. 

\subsection{Training}

\mname can be trained with reinforcement learning (RL) or supervised learning (SL). The RL mode is most natural for handling ARC tasks: the program $p$ synthesized by the Generator is applied to the query raster and returns an output raster, which is then sent to the oracle. The oracle deems it correct or not and that response determines the value of the reward  (1 or 0, respectively), which is then used to update the Generator. In this mode, we rely on the REINFORCE algorithm~\cite{Williams:92,SuttonRL}.  

Unfortunately, the a priori odds for a generated program to solve the given task are minuscule. As a result, training \mname only with RL is usually inefficient, especially in the early stages, when the generated programs are almost entirely random: most episodes lead to no reward and, consequently, no updates of \mname's parameters. This motivates considering the SL mode, in which we assume that the correct program (target) is known. This allows us to directly confront the actions of the Generator (i.e. the AST nodes it produces) with the target program node-by-node, and apply a loss function that rewards choosing the right symbols at individual nodes and penalizes the incorrect choices. In SL, every training episode produces non-zero updates for the model's parameters (unless the target program has been perfectly generated).  

In general, the specific program used as the target in this scenario will be \emph{one of} many programs that implement the input-output mapping required by the demonstrations of the presented task (see Sec.~\ref{sec:arc}). Deterministic models are fundamentally incapable of realizing one-to-many mappings, and the variational layer described in Sec.~\ref{sec:solver} is meant to address this limitation. Upon the (unlikely in practice) perfect convergence of \mname's training, we expect the \emph{deterministic} output of the Solver (corresponding to $z_\mu$) to abstractly represent the common semantic of all programs that solve the presented task, and the \emph{variational} layer to sample the latents that cause the Generator to produce concrete programs \emph{with that very semantics}. 

The prerequisite for the SL mode is the availability of target programs; as those are not given in the ARC benchmark, we devise a method for producing them online during training, presented in the next section.

\subsection{Learning from mistakes}

The programs produced by the Generator are \emph{syntactically correct by construction}. Barring occasional run-time errors (e.g., applying a function to an empty list), a generated program will thus always produce \emph{some} output raster for a given input raster; we refer to it as \emph{response}. By applying such a program $p$ (and arguably \emph{any} syntactically correct program with the \emph{Region} $\rightarrow$ \emph{Region} signature) to the list $I$ of input rasters of some task $T$, we obtain the corresponding list of responses $O$. We observe that the resulting raster pairs made of the elements of $I$ and $O$ \emph{can be considered as another ARC task} $T'$, to which $p$ is the solution (usually \emph{one of} possible solutions, to be precise). The resulting pair $(T',p)$ forms thus a complete example that can be used to train \mname in SL mode, as explained in the previous section, where $T'$ is presented to \mname and $p$ is the target program. 

This observation allows us to \emph{learn from mistakes}: whenever the Generator produces a program $p$ that fails the presented training task $T$, we pair it with the task $T'$ created in the above way, add the \emph{synthetic task} $(T',p)$ formed in this way to the working collection $S$ of \emph{solved tasks}, and subsequently use them for supervised learning. Crucially, we expect $T'$ to be on average easier than $T$, and thus provide the training process with a valuable `learning gradient'. By doing so, we intend to help the model make progress in the early stages of training, when its capabilities fall far behind the difficulty of the ARC tasks. 

To model the many-to-many relation between tasks and programs, we implement $S$ as a relational database to facilitate the retrieval of all programs (known to date) that solve a given task, and vice versa --- of all tasks solved by a given program. 
We disallow duplicates in $S$.

\begin{figure}[t!]
    \centering
    \includegraphics[width=1\linewidth]{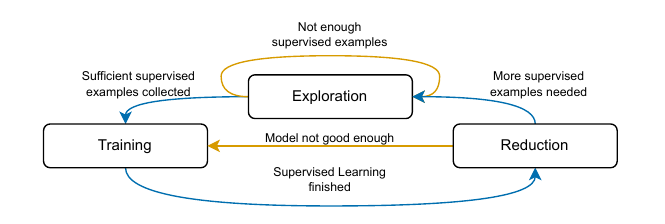}
    \caption{The state diagram of \mname's training, including learning from mistakes.  
    }
    \label{fig:phases}
\vspace{-3mm}
\end{figure}

We start with $S=\emptyset$ and $L$ filled with the original ARC tasks, and stage learning into cycles of \emph{Exploration}, \emph{Training}, and \emph{Reduction} phases (Fig.~\ref{fig:phases}).

\vspace{1mm}\noindent\textbf{Exploration}. The purpose of this phase is to provide synthetic training examples needed in subsequent phases. A random task $T$ is drawn from $L$ and the Generator is queried on it. If the generated program $p$ solves $T$, the pair $(T,p)$ is added to $S$. 
Otherwise, it is checked whether the responses produced by $p$ meet basic criteria of nontriviality, i.e. are non-empty and depend on the input rasters (i.e. responses vary by demonstration). If $T'$ passes this test, $(T',p)$ is added to the $S$ and $L$. 
This continues until enough new tasks have been added to $S$.

\vspace{1mm}\noindent\textbf{Training} consists in applying SL to a random subset of tasks drawn from $S$. The execution of this phase ends after iterating through all training examples in the drawn subset.

\vspace{1mm}\noindent\textbf{Reduction} starts with selecting a subset with known solutions (programs). Then, those tasks are grouped by solutions; a group of tasks with the same solution forms a \emph{category}. Next, $n$ categories are drawn at random. Finally, $k$ tasks are drawn for each category. The tasks selected in this way form a working subset $L' \subset L$.

In the next step, \mname is evaluated on $L'$.
If the program produced by the Generator solves a given task from $L'$, the task is marked as \emph{learned} and is removed from $S$ (if present in $S$). Otherwise, the task is marked as \emph{not learned} and is added to $S$ (if not present in $S$)\footnote{In this way, we allow for re-evaluation of tasks marked previously as \emph{learned} and removed from $S$.}. Finally, the results are grouped according to the category from which the tasks come and the average of solved tasks within each of them is calculated. If \mname reaches the average value of solved categories above the set threshold in the last iterations and stagnation occurs, we switch to the Exploration phase; otherwise, to the Training phase.

\section{Related work}\label{sec:related}

\mname engages programs to process and interpret the input data, and thus bears similarity to several past works on neurosymbolic systems, of which we review only the most prominent ones. In synthesizing programs in response to input (here: task), \mname resembles the Neuro-Symbolic Concept Learner (NSCL, \cite{Mao_2019}). NSCL was designed to solve Visual Query Answering tasks~\cite{JohnsonCLEVR} and learned to parameterize a semantic parser that translated a natural language query about scene content to a DSL program which, once executed, produced the answer to the question. The system was trained with RL. Interestingly, NSCL's DSL was implemented in a differentiable fashion, which allowed it to inform its perception subnetwork in training. 

In using a program synthesizer to produce new tasks, rather than only to solve the presented tasks, \mname bears some similarity to the DreamCoder \cite{Ellis_2021}. DreamCoder's training proceeds in cycles comprising \emph{wake}, \emph{dreaming}, and \emph{abstraction} phases which realize respectively solving the original problems, training the model on `replays' of the original problems and on `fantasies' (synthetic problems), and refactorization of the body of synthesized programs. The last phase involves identification of the often-occurring and useful snippets of programs, followed by encapsulating them as new functions in the DSL, which is meant to facilitate `climbing the abstraction ladder'. The DreamCoder was shown to achieve impressive capabilities on a wide corpus of problems, ranging from typical program synthesis to symbolic regression to interpretation of visual scenes. For a thorough review of other systems of this type, the reader is referred to \cite{Chaudhuri_2021}.

\section{Experimental evaluation}\label{sec:experiment}

In the following experiment, we examine \mname's capacity to provide itself with a `reflexive learning gradient', meant as continuous supply of synthetic tasks at the level of difficulty that facilitates further improvement. Therefore, we focus on the dynamics of the learning process. 

\vspace{1mm}\noindent\textbf{Setup}. To ensure a sufficiently large pool of training examples, each Exploration phase lasts until the set $S$ contains at least 8192 tasks and 32 unique solutions. For the Reduction phase, we set the number of categories to be drawn for $L'$ to $n=64$, the number $k$ of tasks to be drawn from each category to 32, the solving threshold of 30\%, and the number of stagnation iterations to 10. Moreover, generated programs are limited to a maximum number of nodes of 64, a maximum depth of 8 and at most of 2 nestings of higher-order functions (each nesting is considered as a separate program and is also required to meet the above limits).

\vspace{1mm}\noindent\textbf{Metrics}. 
The primary metric is the percentage of tasks solved from the testing subset of the original ARC (\textbf{RateARC}). However, because of the difficulty of this corpus, this metric is very coarse. To assess the progress in a more fine-grained way, we prepare a collection of 183{,}282 synthetic tasks by collecting them from several past runs of the method. This collection is fixed; the percentage of tasks solved from that collection will be referred to as \textbf{RateSynth}.

\begin{table}[t!]
\centering
\caption{RateSynth and RateARC in consecutive training cycles.}
\label{tab:synthrate}
\begin{tabular}{@{\hspace{1em}}r@{\hspace{1em}}c@{\hspace{0.5em}}c@{\hspace{0.5em}}c@{\hspace{0.5em}}c@{\hspace{0.5em}}c@{\hspace{1em}}}
\toprule
Cycle                & 1        & 2        & 3        & 4        & 5        \\ \midrule
RateSynth            & 1.72\% & 4.23\% & 9.81\% & 13.95\% & 21.66\% \\
RateARC            & 1.00\% & 0.25\% & 0.75\% & 1.50\% & 2.00\% \\ \bottomrule
\end{tabular}
\vspace{-5mm}
\end{table}


\vspace{1mm}\noindent\textbf{Results}. 
Table \ref{tab:synthrate} presents the metrics at the completion of consecutive training cycles (cf. Fig.\ \ref{fig:phases}). RateSynth monotonously increases over time, indicating steady progress of \mname's capacity of solving tasks. This positively impacts the RateARC, which achieves the all-high value of 2\% at the end of the run, suggesting that the skills learned from the synthetic, easier tasks translate into more effective solving of the harder original ARC tasks.  

\begin{figure}[t!]
    \centering
    \includegraphics[width=1\linewidth]{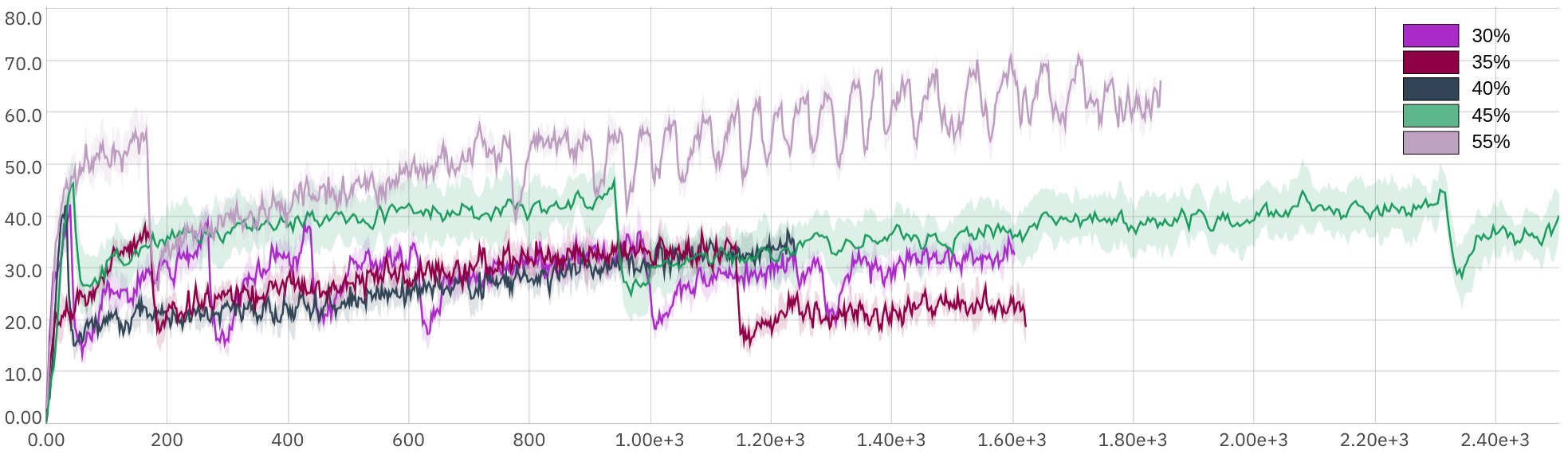}
    \caption{Solving rate after each Reduction phase for \mname runs with different exploration and reduction parameters. 
    The graph shows runs with different thresholds of the solved problem rate that trigger the transition from Reduction to Exploration.}
    \label{fig:evalrate}
\end{figure}

\begin{figure}[t!]
    \centering
    \includegraphics[width=0.7\linewidth]{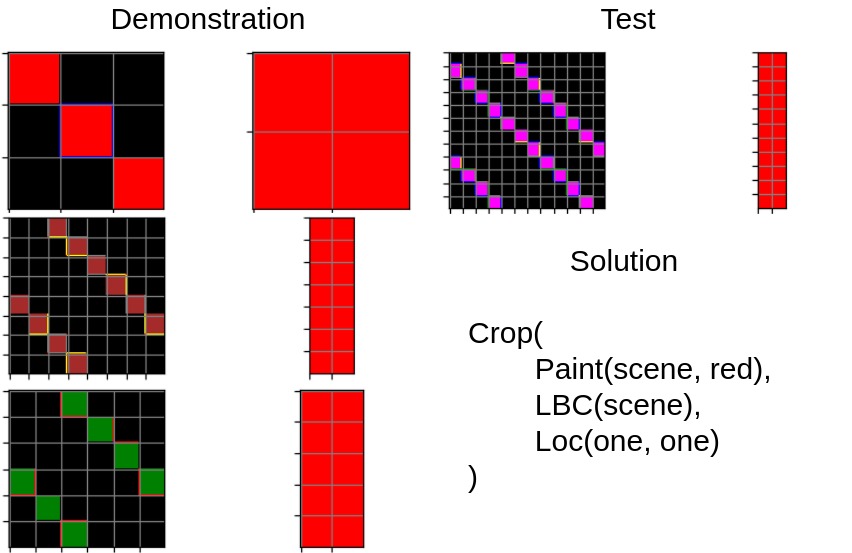}\vspace{6mm}
    \includegraphics[width=0.7\linewidth]{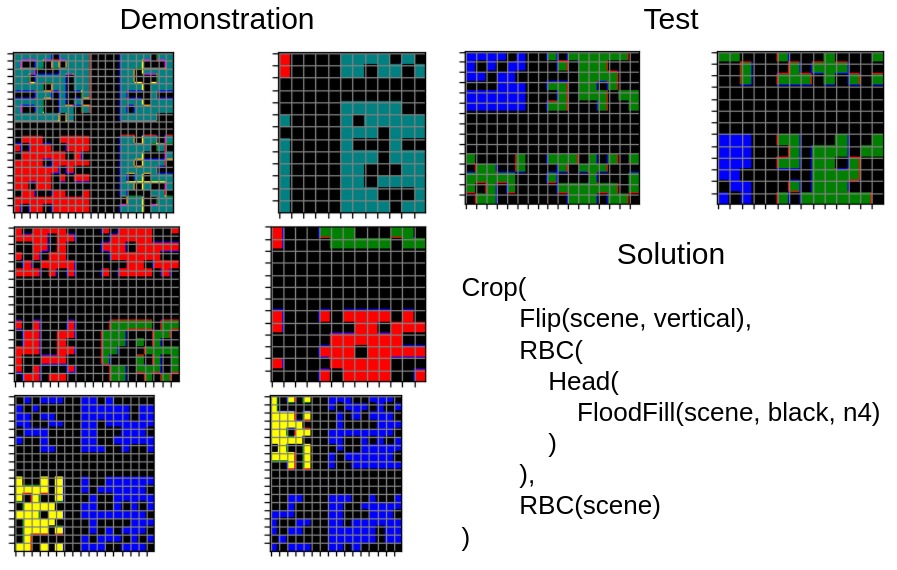}
    \caption{Examples of (task, program) pairs synthesized by the model.}
    \label{fig:syntharc}
\end{figure}

Figure \ref{fig:evalrate} shows the percentage of tasks solved estimated from a random sample drawn from the current $S$. Because $S$ varies dynamically along training, this quantity is not objective, yet illustrates the dynamics of training. The sudden drops in performance occur right after the completion of the Exploration phase, which augments $S$ with new tasks that the method cannot yet solve. Figure~\ref{fig:syntharc} shows examples of generated tasks with solutions, i.e. the $(T,p)$ pairs added to $S$ during training.

Table \ref{tab:progressrate} provides yet another perspective: the performance of the snapshots of \mname trained for a given number of cycles (in rows) on $S$s collected in particular cycles (in columns).\footnote{Calculated off-line, after the completion of the run.} Similarly to previous results, the table demonstrates overall consistent improvement of the model's performance. Furthermore, the metric decreases only twice within columns, which suggests that losing the capacity to solve tasks that were solved in the past occurs only occasionally. 

\begin{table}[t!]
\centering
\caption{Performance of the snapshot of the model from a given cycle (row) on the synthetic examples collected in a given cycle (column). For instance, the model preserved in cycle 2 solves 0.47\% of synthetic tasks created in cycle 3. 
}
\label{tab:progressrate}
\begin{tabular}{@{\hspace{1em}}l@{\hspace{0.5em}}c@{\hspace{0.5em}}r@{\hspace{0.5em}}r@{\hspace{0.5em}}r@{\hspace{0.5em}}r@{\hspace{0.5em}}r@{\hspace{1em}}}
\toprule
\multicolumn{1}{l}{}  & \multicolumn{1}{l}{} & \multicolumn{5}{c}{Synthetic task set $S$ from cycle}                    \\ \midrule
\multicolumn{1}{l}{} & \mname's snapshot from cycle & \multicolumn{1}{c}{1} & \multicolumn{1}{c}{2} & \multicolumn{1}{c}{3} & \multicolumn{1}{c}{4} & \multicolumn{1}{c}{5} \\ \midrule
    & 1 & 35.08\% & 0.56\%  & 0.05\%  & 0.00\%  & 0.01\%  \\
    & 2 & 33.35\% & 29.68\% & 0.47\%  & 0.11\%  & 0.11\%  \\
    & 3 & 43.06\% & 30.44\% & 34.18\% & 1.35\%  & 0.65\%  \\
    & 4 & 43.53\% & 28.11\% & 30.33\% & 24.33\% & 1.84\%  \\
    & 5 & 49.03\% & 31.22\% & 31.40\% & 30.74\% & 24.74\% \\ \bottomrule
\end{tabular}
\vspace{-5mm}
\end{table}

\section{Conclusions and future work}\label{sec:conclusion}

This study summarized our preliminary findings on \mname and illustrated its overall capacity to provide itself with a learning gradient. Crucially, the generative aspect of this architecture, combined with expressing candidate solutions in a DSL, allows the method to obtain concrete target DSL programs and so gradually transform an unsupervised learning problem into a supervised one. As evidenced by the experiment, supervised learning facilitated in this way provides more informative learning guidance than reinforcement learning.  

The modularity of the proposed architecture allows the model to be adapted for other types of data. In particular, the Solver and Generator modules are independent of the input data type, while the only type-specific module is Perception. Future work will include applying the approach to other benchmarks in different domains, developing alternative interchangeable DSLs, transferring abstraction and reasoning knowledge between datasets, and prioritizing the search in the solution space to solve the original ARC tasks.

\section*{Acknowledgements} 
This research was supported by TAILOR, a project funded by EU Horizon 2020 research and innovation program under GA No. 952215, by the statutory funds of Poznan University of Technology and the Polish Ministry of Education and Science grant no. 0311/SBAD/0726.

\clearpage

\section*{Appendix}\label{app:dsl}
\subsection*{Specification of the DSL}

\begin{table}[h!]
\centering
\caption{The list of types available in the DSL.}
\begin{tabular}{@{}p{0.2\linewidth}p{0.8\linewidth}@{}}
\label{tab:dsltype}\\
\toprule
Name           & Description                                                                                   \\ \midrule
Arithmetic     & An abstract type that implements basic arithmetic operations such as addition and subtraction \\
Bool           & Logical type; accepts True/False values                                                       \\
Color          & Refers to the categorical value of pixels. It can take one of ten values                      \\
Comparable & An abstract type that implements basic operations that allow objects to be compared with each other               \\
Int            & Simple integer type                                                                           \\
Loc            & A location consisting of two integers                                                         \\
Connectivity   & The type of neighborhood used by the FloodFill operation; possible values are n4 and n8       \\
Direction  & The type of direction used by the Rotate operation; possible values are cw (clockwise) and cww (counterclockwise) \\
Orientation    & The type of direction used by the Flip operation; possible values are vertical and horizontal \\
Region         & An object representing any list of pixels and their colors                                    \\
List{[}T{]}    & A generic type representing a list of objects of a compatible type                            \\
Pair{[}T, L{]} & A generic type representing a pair of objects of a compatible type                            \\* \bottomrule
\end{tabular}
\end{table}

\begin{table}[h!]
\centering
\caption{The list of predefined symbol keys available in the DSL. The keys \emph{Zero}, \emph{One}, \emph{Horizontal}, \emph{Vertical}, \emph{N4}, \emph{N8}, \emph{Cw}, and \emph{Ccw} represent constant values and are always present in a Workspace. \emph{Colors} are only available if they appear within a given task. \emph{Scene} is a key relative to a specific pair of demonstrations. \emph{FunctionalInput} is a special key used by higher-order functions. 
}
\label{tab:dslsymbols}
\begin{tabular}{@{}p{0.15\linewidth}p{0.2\linewidth}p{0.7\linewidth}@{}}
\toprule
Name       & Type         & Description                                          \\ \midrule
Zero       & Int          & A constant `0`                                       \\
One        & Int          & A constant `1`                                       \\
Horizontal & Orientation  & A categorical value used for the Flip operation      \\
Vertical   & Orientation  & A categorical value used for the Flip operation      \\
N4         & Connectivity & A categorical value used for the FloodFill operation \\
N8         & Connectivity & A categorical value used for the FloodFill operation \\
Cw         & Direction    & A categorical value used for the Rotate operation    \\
Ccw        & Direction    & A categorical value used for the Rotate operation    \\
Black      & Color        & A categorical value of color from ARC                \\
Blue       & Color        & A categorical value of color from ARC                \\
Red        & Color        & A categorical value of color from ARC                \\
Green      & Color        & A categorical value of color from ARC                \\
Yellow     & Color        & A categorical value of color from ARC                \\
Grey       & Color        & A categorical value of color from ARC                \\
Fuchsia    & Color        & A categorical value of color from ARC                \\
Orange     & Color        & A categorical value of color from ARC                \\
Teal       & Color        & A categorical value of color from ARC                \\
Brown      & Color        & A categorical value of color from ARC                \\
Scene            & Region & A Region representing input raster from an input-output demonstration pair                      \\
Functional Input & -      & A special key available only during execution/generation of functional operation subprogram \\ \bottomrule
\end{tabular}
\end{table}

\begin{landscape}
\begin{longtable}[c]{@{}p{0.12\linewidth}p{0.33\linewidth}p{0.55\linewidth}@{}}
\caption{Definitions of the operations available in the DSL.}
\label{tab:dslops}\\
\toprule
Name &
  Signature &
  Description \\* \midrule
\endhead
\bottomrule
\endfoot
\endlastfoot
Add &
  (A: Arithmetic, A: Arithmetic) -\textgreater A: Arithmetic &
  Adds two objects of the same type inheriting from Arithmetic \\
Area &
  (Region) -\textgreater Int &
  Calculates the surface area of a region \\
Crop &
  (Region, Loc, Loc) -\textgreater Region &
  Cuts a subregion based on the top left and bottom right vertices \\
Deduplicate &
  (List{[}A{]}) -\textgreater List{[}A{]} &
  Removes duplicates from the list \\
Diff &
  (List{[}A{]}, List{[}A{]}) -\textgreater List{[}A{]} &
  Performs a difference operation on two lists \\
Draw &
  (Region, Union{[}Region, List{[}Region{]}{]}) -\textgreater Region &
  Overlays a Region or list of Regions on the given input region \\
Equals &
  (A, A) -\textgreater Bool &
  Compares two objects of the same type \\
Filter &
  (List{[}A{]}, (A-\textgreater{}Bool)) -\textgreater List{[}A{]} &
  Filters the list of input objects based on the result of the subroutine run on each input element \\
First &
  (Pair{[}A, Type{]}) -\textgreater A &
  Gets the first element of the input pair \\
Flip &
  (Region, Orientation) -\textgreater Region &
  Performs a vertical or horizontal flip of the input region \\
FloodFill &
  (Region, Color, Connectivity) -\textgreater List{[}Region{]} &
  Performs segmentation of the input Region using the Flood Fill algorithm and the background color and neighborhood type (n4 or n8) \\
GroupBy &
  (List{[}A{]}, (A-\textgreater{}B)) -\textgreater List{[}Pair{[}B, List{[}A{]}{]}{]} &
  Groups objects from the input list based on the results of the subroutine \\
Head &
  (List{[}A{]}) -\textgreater A &
  Gets the first element of the input list \\
Height &
  (Region) -\textgreater Int &
  Returns the height of the region \\
Intersection &
  (List{[}A{]}, List{[}A{]}) -\textgreater List{[}A{]} &
  Returns the intersection of two lists \\
LBC &
  (Region) -\textgreater Loc &
  Returns the location of the left bottom corner \\
LTC &
  (Region) -\textgreater Loc &
  Returns the location of the left top corner \\
Len &
  (List{[}Type{]}) -\textgreater Int &
  Returns the length of the input list \\
Line &
  (Loc, Loc, Color) -\textgreater Region &
  Creates a single-color Region that represents a line between two points \\
Loc &
  (Int, Int) -\textgreater Loc &
  Location object constructor \\
Map &
  (List{[}A{]}, (A-\textgreater{}B)) -\textgreater List{[}B{]} &
  Performs the transformation operation of each element of the input list using a subroutine \\
MostCommon &
  (List{[}A{]}, (A-\textgreater{}B)) -\textgreater B &
  Returns the most frequently occurring object from the input list based on the value returned by the subroutine applied to the list elements \\
Neg &
  (Union{[}A, B: Bool{]}) -\textgreater Union{[}A, B: Bool{]} &
  Performs a negation operation on the input object \\
Paint &
  (Region, Color) -\textgreater Region &
  Colors the input region a solid color \\
Pair &
  (A, B) -\textgreater Pair{[}A, B{]} &
  The constructor of an object of type Pair \\
Pixels &
  (Region) -\textgreater List{[}Region{]} &
  Returns a list of pixels of the input region \\
RBC &
  (Region) -\textgreater Loc &
  Returns the location of the right bottom corner \\
RTC &
  (Region) -\textgreater Loc &
  Returns the location of the right top corner \\
Rect &
  (Loc, Loc, Color) -\textgreater Region &
  Creates a single-color Region representing a rectangle bounded by the input locations \\
Reverse &
  (List{[}A{]}) -\textgreater List{[}A{]} &
  Reverses the input list \\
Rotate &
  (Region, Direction) -\textgreater Region &
  Rotates the Input Region clockwise or counterclockwise \\
Scale &
  (Region, Union{[}Int, Pair{[}Int, Int{]}{]}) -\textgreater Region &
  Scales the input Region according to the integer argument \\
Second &
  (Pair{[}Type, A{]}) -\textgreater A &
  Returns the second element of the input pair \\
Shift &
  (Region, Union{[}Loc, Pair{[}Int, Int{]}{]}) -\textgreater Region &
  Shifts the input Region by an integer argument \\
Sort &
  (List{[}A{]}, (A-\textgreater{}Comparable) -\textgreater List{[}A{]} &
  Sorts the input list based on the result of the subroutine applied to the list \\
Sub &
  (A: Arithmetic, A: Arithmetic) -\textgreater A: Arithmetic &
  Subtracts two objects of the same type inheriting from Arithmetic \\
Tail &
  (List{[}A{]}) -\textgreater A &
  Gets the last element of the input list \\
Union &
  (List{[}A{]}, List{[}A{]}) -\textgreater List{[}A{]} &
  Returns the sum of two input lists \\
Width &
  (Region) -\textgreater Int &
  Returns the width of the region \\
Zip &
  (List{[}A{]}, List{[}B{]}) -\textgreater List{[}Pair{[}A, B{]}{]} &
  Creates a list of pairs of corresponding elements in the input lists \\* \bottomrule
\end{longtable}
\end{landscape}

\bibliographystyle{unsrtnat}
\bibliography{main}

\end{document}